\documentclass[10pt, conference]{IEEEtran}

\ifCLASSOPTIONcompsoc
  \usepackage[nocompress]{cite}
\else
  \usepackage{cite}
\fi

\ifCLASSINFOpdf
\else
\fi

\usepackage{graphicx}

\usepackage{url}
\usepackage{cite}
\usepackage{amssymb}
\usepackage{amsmath}
\usepackage{cases}

\usepackage{color}
\usepackage{comment}
\usepackage[linesnumbered,lined,boxed,commentsnumbered,ruled,longend]{algorithm2e}

\usepackage[subrefformat=parens,labelformat=parens]{subfig}
\usepackage{balance}

\IEEEoverridecommandlockouts\IEEEpubid{\makebox[\columnwidth]{ 978-1-6654-8234-9/22/\$31.00 $\copyright$2022 IEEE \hfill}\hspace{\columnsep}\makebox[\columnwidth]{ }}

\begin{document}

\title{Boost Decentralized Federated Learning in Vehicular Networks by Diversifying Data Sources}

\author{\IEEEauthorblockN{Dongyuan Su\IEEEauthorrefmark{4}, Yipeng Zhou\IEEEauthorrefmark{3} and Laizhong Cui\IEEEauthorrefmark{4} \IEEEauthorrefmark{1}}

\IEEEauthorblockA{\IEEEauthorrefmark{4}College of Computer Science and Software Engineering, Shenzhen University, Shenzhen, China\\
Email: sudongyuan2019@email.szu.edu.cn, cuilz@szu.edu.cn}

\IEEEauthorblockA{\IEEEauthorrefmark{3}School  of Computing, Faculty of Science and Engineering, Macquarie University, Sydney, Australia\\Email: yipeng.zhou@mq.edu.au}

\thanks{This work has been partially supported by National Key R\&D Program of China under Grant No.2018YFB1800302, Shenzhen Science and Technology Program under Grant No. RCYX20200714114645048, and No. JCYJ20190808142207420, and Shenzhen Fundamental Research Program under Grant No. 20200814105901001.}
\IEEEauthorblockA{
		\thanks{\textit{(Corresponding author: Laizhong Cui)}
	}
}
}

\maketitle

%\IEEEtitleabstractindextext{%
\begin{abstract}
Recently, federated learning (FL) has received intensive research because of its ability in  preserving data privacy for scattered clients to collaboratively train machine learning models. Commonly, a parameter server (PS) is deployed for aggregating model parameters contributed by different clients.  Decentralized federated learning (DFL) is upgraded from FL which allows  clients to aggregate model parameters with their neighbours directly.  DFL is particularly feasible for vehicular networks as vehicles communicate with each other in a vehicle-to-vehicle (V2V) manner.  However, due to the restrictions of vehicle routes and communication distances, it is hard for individual vehicles to sufficiently exchange models with others.  Data sources contributing to models on individual vehicles may not diversified enough resulting in poor model accuracy. To address this problem, we propose the DFL-DDS (DFL with diversified Data Sources) algorithm to diversify data sources in DFL. Specifically, each vehicle  maintains a state vector to record the contribution weight of each data source to its model. 
The Kullback–Leibler (KL) divergence is adopted to measure the  diversity of a state vector. 
To boost the convergence of DFL, a vehicle tunes the aggregation weight of each data source by minimizing the KL divergence of its state vector, and its effectiveness in diversifying data sources can be theoretically proved.
Finally, the superiority of DFL-DDS is evaluated by extensive experiments (with  MNIST and CIFAR-10 datasets) which demonstrate that DFL-DDS can accelerate the convergence of DFL and improve the model accuracy significantly compared with state-of-the-art baselines. 
\end{abstract}

%}

\section{Introduction}

With the proliferation of intelligent vehicles, vehicles have accumulated abundant datasets via equipped LiDAR, RIDAR, camera and sensors \cite{Liang_2019}. Advanced machine learning models can be obtained by exploiting datasets on vehicles to  improve road safety, object recognition accuracy, etc \cite{UnlabeldDataFL}. 
However, exploiting datasets on vehicles for model training confronts at least two challenges: First, accessing raw datasets residing on vehicles inevitably invades data privacy because these datasets probably contain sensitive and confidential information; Second, the communication cost may be very heavy by extensively collecting raw datasets from vehicles. Thanks to the recent advances of  federated learning (FL),  these challenges can be potentially solved. 

To mitigate the rising concerns on data privacy leakage,  FL conducts model training without touching original data samples stored on scattered clients \cite{FedAvg}. In each round of FL global iteration (a.k.a epoch), a parameter server (PS) is responsible for aggregating model parameters collected from participating clients who conduct local iterations with local datasets and then disclose their model parameters to the PS. Global iterations will be conducted for multiple times by involving different clients until the trained model finally converges \cite{FedAvg}.

To avoid the single point failure  and alleviate the communication bottleneck of the  PS in FL \cite{Central_bottleneck}, decentralized FL (DFL) is devised which allows each client to aggregate model parameters gathered from its neighbours directly. According to \cite{TimeVaryingGraph, Savazzi_20}, clients weigh models from neighbours based on their connection degrees or  sample sizes  for model aggregation. The convergence of DFL has been proved in \cite{DFL_converge,zhong2021p} assuming that the network topology formed by clients is connected. DFL is particularly feasible for vehicular networks provided that vehicles as learning clients communicate with each other in a vehicle-to-vehicle (V2V) manner \cite{lyl_2020}. 

\begin{figure}
\centering
\includegraphics[width=0.8\linewidth]{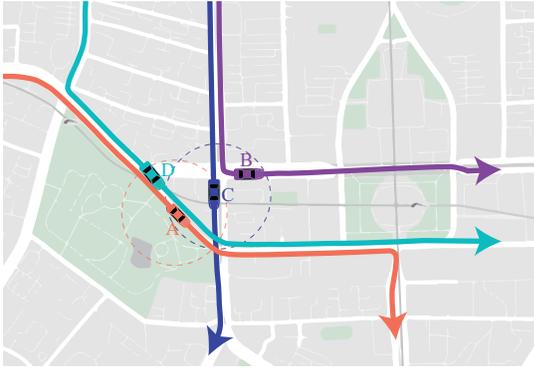}
\caption{A simple example to show the model exchange via an intermediate vehicle in DFL.}
\label{Fig:Case}
\end{figure}

Despite the feasibility of DFL, it is highly possible that individual vehicles \footnote{We use vehicles exchangeably with clients in our paper. }  fail to attain satisfactory model accuracy  due to restrictions of their routes and communication distances \cite{Deveaux2020}.  To illustrate this point, we show a concrete example  in Fig. \ref{Fig:Case} by using real traces extracted from DiDi \cite{DiDi}. In Fig. \ref{Fig:Case}, there are four vehicles, namely A, B, C and D.  The route of vehicle A has no overlap with the route of vehicle B though their information can be exchanged via intermediate vehicle C. In other words, data sources contributing to model aggregation on vehicle A are not sufficiently diversified  given that vehicle A cannot directly communicate with vehicle B. The problem is even worse in a real large-scale vehicular network when vehicles need to exchange model parameters  via multiple hops through intermediate vehicles. As a consequence, the convergence of DFL in vehicular networks could be very slow.

Diversifying data sources in FL has been explored by existing works. It has been regarded as an important data related factor affecting model performance in \cite{LiAnran}, which proposed a diversity-driven client selection strategy. However, a central server is needed to collect the noisy data sketch of each client to compute the data similarity between two clients. In \cite{D2D_offloading}, a joint sampling and data offloading algorithm is proposed to improve  model performance, where the sampling method is based on the utility of device data distributions. A D2D (device-to-device) offloading scheme is used to diversify sampled devices, though D2D offloading between neighboring devices is unsuitable in many privacy-preserving tasks. In \cite{SkewAware}, training workers are assumed to be connected with all data sources, which can ensure that the dataset collected by each training worker is of large diversity. 
Unfortunately, these methods cannot be leveraged to support DFL in vehicular networks in that they failed to  consider the fully decentralized V2V communication manner in vehicular networks.

To boost DFL by diversifying data sources in vehicular networks, we propose the DFL with diversified data sources (DFL-DDS) algorithm. First of all, we implement the distributed learning algorithm designed in \cite{TimeVaryingGraph} without considering data source diversity to conduct a simulation study. Our study unveils that unlucky vehicles will encounter difficulties to diversify their data sources for model aggregation. Consequently,  
their model accuracy  is much lower than that of other vehicles. 
Inspired by our simulation study, we design DFL-DDS as below. Each vehicle in DFL-DDS maintains a state vector to record the contribution weight of each vehicle in the system to its model, which will be exchanged between vehicles along with the exchange of model parameters. 
Formally, we  employ the KL divergence to measure the diversity of a state vector, which not only quantifies data source diversity but also accommodates heterogeneous dataset sizes from different vehicles. Prior to aggregating models, each vehicle generates the aggregation weight for each data source by minimizing the KL divergence of its aggregated state vector, and its effectiveness in diversifying data sources can be theoretically illustrated. To validate that higher model accuracy can be attained with more diversified data sources, we conduct extensive experiments to evaluate our algorithm by using standard  MNIST and CIFAR-10 datasets. Experimental results demonstrate superb performance achieved by DFL-DDS in comparison with competitive baselines.

The rest of the paper is organized as follows. Relevant works and our contribution compared to these works are discussed in Sec. \ref{Sec:Related}. Preliminary knowledge of the FedAvg and Decentralized FedAvg algorithms are introduced in Sec. \ref{Sec:Pre}. The simulation study showing the deficiency of existing algorithms is presented in Sec. \ref{Sec:Mea}. The DFL-DDS algorithm and its analysis are elaborated in  Sec. \ref{Sec:Algorithm}. Experiments are conducted and results are reported in Sec. \ref{Sec:Experiment}. Lastly, we conclude our work and envision our future work in Sec. \ref{Sec:Con}.

\section{Related Work}
\label{Sec:Related}

In this section, we briefly discuss related works from two aspects: federated learning and learning in vehicular networks, and state our contribution in light of  existing works.  

\subsection{Federated Learning}

Federated learning was originally proposed by \cite{FedAvg} to preserve the privacy of  data distributed on decentralized clients during model training. FedAvg and FedSGD \cite{FedAvg, jianmin2016revisiting} were proposed as two most fundamental model averaging algorithms in FL. They can complete model training by exchanging model parameters between the PS and clients without accessing original data samples.

Since its inception, FL has received intensive research efforts, and has been applied in various systems.
To fit in different applications, variants of FedAvg/FedSGD were devised to accelerate  FL.  Khalil \emph{et. al.} extended FedAvg  by introducing a new sampling and aggregation strategy to improve the convergence speed when training a federated recommendation model \cite{FedFast}. In \cite{chen2020fedhealth}, Chen \emph{et. al.} proposed a federated transfer learning framework for wearable healthcare to achieve personalized model learning. In \cite{Yu-2021}, Yu \emph{et. al.} proposed a neural-structure-aware resource management approach with module-based FL framework to improve the resource efficiency. In \cite{nishio2019client}, Nishio \emph{et. al.} proposed algorithms to actively manage clients based on their resource conditions to accelerate FL procedure.

Regardless of application scenarios, the data diversity is always an important factor influencing the accuracy of FL. In FL, data samples are collected and privately maintained by distributed clients, which makes data sample selection and the estimation of data sample quality difficult.  
In view of this difficulty, research efforts have been dedicated to diversifying data sources to improve model accuracy. 
In \cite{SkewAware}, Pu \emph{et. al.} proposed an efficient online data scheduling policy to alleviate the data skew issue caused by the capacity heterogeneity of training workers from the long-term perspective. 
In \cite{D2D_offloading}, Wang \emph{et. al.} proposed a joint sampling and data offloading optimization problem subject to constraints on the network topology and device capacities to improve FL training accuracy. Considering the similarity of datasets collected across devices, a similarity-aware data offloading strategy was proposed in a distributed manner to optimize the data dissimilarity of each device.
In \cite{LiAnran},  Li \emph{et. al.} reported that a low content diversity can lead to severe model accuracy deterioration based on trace driven analysis. A client sampling method jointly considering the statistical homogeneity and content diversity was proposed to improve model accuracy.

However, all these mentioned works trying to improve the diversity of data sources  failed to consider the decentralized FL scenario. Due to the lack of a centralized PS in DFL, it is rather difficult to collect necessary information to  diversify data sources from a global perspective.
In \cite{Savazzi_20}, FedAvg  was extended to a DFL algorithm to enable learning in arbitrarily large scale networks. Same as FedAvg, the aggregation weight is proportional to the sample population failing to diversify data sources.
In \cite{SegmentedGossip}, Hu \emph{et. al.} proposed a segmented gossip aggregation approach allowing each worker to pull different model segments from chosen peers.
The objective is to optimize the utilization of the bandwidth resources between workers requiring high connectivity among workers. 
In \cite{TimeVaryingGraph},  a broadcast-based algorithm was developed to minimize the sum of convex functions scatted throughout a collection of distributed nodes. 
The aggregation weight of each model is merely  determined by the in-degree and out-degree of each worker, which cannot effectively  diversify data sources.

\subsection{Learning in Vehicular Networks}
To accommodate  intelligent applications on vehicles, privacy preserved and efficient learning framework applicable for vehicular networks becomes extremely needed. 
It was reported in \cite{elbir2020federated, OpenIssues} 
that FL can benefit many important applications in autonomous driving. 
The works \cite{FVN, ReliableFL, TwoLayerFL} investigated how to  support data- and computation-intensive applications with low communication overhead with a strong privacy guarantee in vehicular networks. 

Nevertheless, it is costly to deploy a PS to conduct centralized model aggregation for FL in large-scale vehicular networks. 
In \cite{samarakoon2018federated}, an RSU (Road Side Unit) based model aggregation method was proposed to conduct FL.
However, pervasively deploying RSUs can be very costly. Thereby, pure V2V solutions are extensively explored by \cite{Luo_ICC_2017, bradai2014vicov, Lin_TMC_17, fan2021high}.
These works indicate the feasibility to implement DFL in vehicular networks by allowing each vehicle to aggregate models collected from neighbours directly. 
In \cite{Entropy, FL_veh},  blockchain-based FL was proposed to enable decentralized FL. Whereas, the FL efficiency of these works cannot be guaranteed.

Based on the above discussion, existing works on DFL in vehicular networks largely overlooked the influence of data source diversity when aggregating models. Consequently, it is hard for them to achieve high model accuracy on every individual vehicles.

\section{Preliminary}
\label{Sec:Pre}

In this section, we briefly explain the FedAvg and Decentralized FedAvg algorithms to facilitate the discussion of our DFL-DDS algorithm presented in the next section.

\subsection{FedAvg}
Without loss of generality, we suppose that $n$ data samples $[(\mathbf{x}_i, y_i)]_{i=1}^{n}$ are distributed among $K$ clients in the FL system. The process to train a machine learning model can be regarded as the process to minimize a loss function. The loss function of a particular sample $i$ can be defined as $f_i(\mathbf{w}) = {l}(\mathbf{x}_i, y_i; \mathbf{w})$, where $\mathbf{w}$ represents model parameters to be learned. The objective  of FL is to minimize the overall loss function defined as 
\begin{equation}
    \label{eq:obj_def}
    \min_{\mathbf{w}} f(\mathbf{w}), \quad\text{where}\quad f(\mathbf{w}) \triangleq \frac{1}{n} \sum_{i=1}^{n} f_i(\mathbf{w}).
\end{equation}

Let $\mathcal{N}_k$ denote the set of data samples owned by client $k$ and $n_k = |\mathcal{N}_k|$. The objective in Eq. \eqref{eq:obj_def} can be rewritten as
\begin{equation}
    f(\mathbf{w}) = \sum_{k=1}^{K} \frac{n_k}{n} F_k(\mathbf{w}), \quad\text{where}\quad F_k(\mathbf{w}) = \frac{1}{n_k} \sum_{i \in \mathcal{N}_k} f_i(\mathbf{w}).
\end{equation}

FederatedAveraging (FedAvg) minimizes the loss function by conducting multiple global iterations (a.k.a epochs) with the assistance of the PS. The learning process consists of the following steps.
\begin{itemize}
    \item The PS randomly and uniformly selects $m$ clients, denoted by $\mathcal{P}_t$, to participate local model update in the $t$-th global iteration. This client selection scheme can ensure the diversity of data sources for model aggregation. 
    \item Each participating client $k$ downloads the latest global model\footnote{The global model in the first global iteration can be randomly initialized by the PS. } from the PS to perform $E$ local updates according to the formula of each local update as below with local data samples. 
        \begin{equation}
        \label{EQ:LocalIter}
            \mathbf{w}^k \leftarrow \mathbf{w}^k - \eta \nabla F_k(\mathbf{w}^k, \mathcal{B}_k), \quad\forall k,
        \end{equation}
        where $\eta$ is the learning rate and $\mathcal{B}_k$ is a batch of samples selected from $\mathcal{N}_k$. Then, the updated model parameters are uploaded to the PS.
    \item The PS aggregates all updated models from participating clients as
        \begin{equation}
        \label{EQ:GlobalIter}
            \mathbf{w}_{t+1} \leftarrow \sum_{k\in \mathcal{P}_t} \alpha_{k,t} \mathbf{w}_{t+1}^k.
        \end{equation}
         Here $\alpha_{k,t}$ represents the weight of participating client $k$ for model aggregation in the $t$-th global iteration. It is usual to set $\alpha_{k,t}$ proportional to $n_k$.
        \item Repeat the above three steps until the model converges. 
\end{itemize}
\subsection{Decentralized FedAvg}
Decentralized FedAvg can be easily extended from FedAvg by enabling each client to collect model parameters from its neighbour clients directly. It is equivalent to implementing the function of the PS on each client. Many existing works can be perceived as variants of decentralized FedAvg such as the sub-gradient push algorithm proposed in \cite{TimeVaryingGraph}. 

Nevertheless, it is tricky to properly set $\alpha_{k, t}$ when deploying DFL in vehicular networks. Considering the constraints of vehicle routes and communication distances, an individual vehicle cannot uniformly and randomly select other vehicles as its neighbours to  diversify data sources contributed to its model training.    Consequently, it becomes a challenging problem  to properly set   $\alpha_{k, t}$ for DFL in vehicular networks. In \cite{TimeVaryingGraph, zhong2021p}, it was proposed to set $\alpha_{k, t}$ according to in-degrees and out-degrees of client $k$, which has ignored the influence of data source diversity on the model accuracy. We will further explore the deficiency to implement such algorithms in vehicular networks with a simulation study in the next section.

\section{Simulation Study}
\label{Sec:Mea}

To visualize the impact of data source diversity on the final model accuracy in DFL, we conduct a simulation study by implementing the subgradient push (SP) algorithm \cite{TimeVaryingGraph} with two public datasets: CIFAR-10 and MNIST.

\subsection{System Model}

We study a vehicular network with $K$ vehicles. Each vehicle owns a dataset $\mathcal{N}_k$. Each vehicle is equipped with certain computation and communication capacity. It can either park or move along roads. We consider a synchronized DFL system. Each vehicle $k$ conducts $E$ local iterations according to Eq.~\eqref{EQ:LocalIter} with its own dataset $\mathcal{N}_k$ and its model $\mathbf{w}^k_t$ in the $t$-th global iteration. When local iterations are completed, all vehicles make efforts to contact  other vehicles in their proximity. Then, two vehicles can exchange model parameters if their distance is within the communication range.  
Let $\mathcal{M}_{k,t}$  denote the set of neighbor vehicles, which can exchange  models with vehicle $k$ for the $t$-th round of model aggregation. Let $\mathcal{P}_{k,t} = \mathcal{M}_{k,t}\cup \{k\}$ denote the set of all vehicles involved in the $t$-th  round of model aggregation on vehicle $k$.  After collecting model parameters from vehicles in $\mathcal{M}_{k,t}$, vehicle $k$ conducts  aggregation according to Eq.~\eqref{EQ:GlobalIter}. With refined  parameters, vehicles proceed to the next round of global iteration. 

To facilitate the understanding of our work, frequently used notations are summarized with brief explanations in Table~\ref{Tab:notation_table}.
\begin{table}
    \centering
    \caption{{\color{black}Frequently used notations}}
    \label{Tab:notation_table}
    \begin{tabular}{cc}
        \hline \hline
        Notation            & Short Explanation \\
        \hline
        $n_k$             &  The number of data samples  owned by vehicle $k$   \\
         $\mathcal{N}_k$         & The local dataset owned by vehicle $k$ \\
        $K$             &  The total number of vehicles         \\
        $E$             &  The number of local updates   \\
        $\eta_t$          & The learning rate             \\
        $\alpha^k_{k',t}$  & The weight of the model from vehicle $k'$ for model \\
         &  aggregation on vehicle $k$ in the $t$-th global iteration \\
        $\mathbf{w}_{t}^k$      & The model of vehicle $k$ in the $t$-th global iteration \\
        $\mathcal{M}_{k,t}$     & The set of vehicles which can communicate with vehicle $k$ \\
        & in the $t$-th global iteration\\
        $\mathcal{P}_{k,t}$     & The set of vehicles involved for model aggregation on  \\
        & vehicle $k$ in the $t$-th global iteration\\
        $\mathbf{s}_{k,t}$      & The state vector on vehicle $k$ in the $t$-th global iteration \\
        $s^k_{k',t}$            & The contribution weight from vehicle $k'$ to vehicle $k$          \\
        $\mathbf{g}$            & The target state vector           \\
        \hline \hline
    \end{tabular}
\end{table}

\subsection{Simulation Implementation}
We implement a simulator to conduct DFL experiments in vehicular networks for our study. 
The SP algorithm introduced in \cite{TimeVaryingGraph} is implemented on each vehicle.
Specifically,  in the $t$-th global iteration, each vehicle $k$ conducts a round of local iteration with its local samples to update the model. Meanwhile, it maintains two intermediate quantities: a vector variable $\mathbf{x}_k(t)$, and  a scalar variable $y_k(t)$. 
In SP, $\mathbf{x}_k(t)$ represents model parameters contributed from client $k$. However, $\mathbf{x}_k(t)$ must be adjusted by the scalar $y_k(t)$ for model aggregation. 
Each vehicle evenly broadcasts two quantities $\mathbf{x}_{k'}(t)/p_{k',t}$ and $y_{k'}(t)/p_{k',t}$ to all neighbour vehicles  $\forall k' \in \mathcal{P}_{k,t}$, where $p_{k',t}= |\mathcal{P}_{k',t}|$, after conducting a local iteration. Then,  $\mathbf{x}_k(t)$ and $y_k(t)$ are utilized to update model parameters. For detailed derivation of $\mathbf{x}_k(t)$ and $y_k(t)$, please refer to \cite{TimeVaryingGraph}. 

We leverage a microscopic traffic simulator SUMO \cite{SUMO2018} to generate grid topology and random topology as the road networks as well as vehicle trajectories. We generate  100 vehicles moving along roads in road networks based on the Manhattan mobility model \cite{Manhattan}. Each vehicle can only communicate to other vehicles within the communication range of 100 meters. Two public datasets, CIFAR-10 with 50k/10k training/test images and MNIST with 60k/10k training/test images, are adopted to conduct our experiments. In each experiment, the training 
samples are distributed to 100 vehicles in a balanced and non-IID manner. In other words,  each vehicle is assigned with 500 training images of CIFAR-10 or 600 training images of MNIST, selected from 2 to 4 out of 10 labels.
Two CNN models are adopted for the image classification tasks. The CNN models for CIFAR-10 and MNIST have 33,834 and  21,840 parameters, respectively.\footnote{These models are from the public code repository at https://github.com/AshwinRJ/Federated-Learning-PyTorch}
We train the CNN models for 3,000 iterations on CIFAR-10 and 300 iterations on MNIST.

\begin{figure}[htb]
 \centering
 \subfloat[CIFAR-10]{
 \label{fig:Accuracy_CDF_CIFAR-10}
 \includegraphics[width=0.37\textwidth]{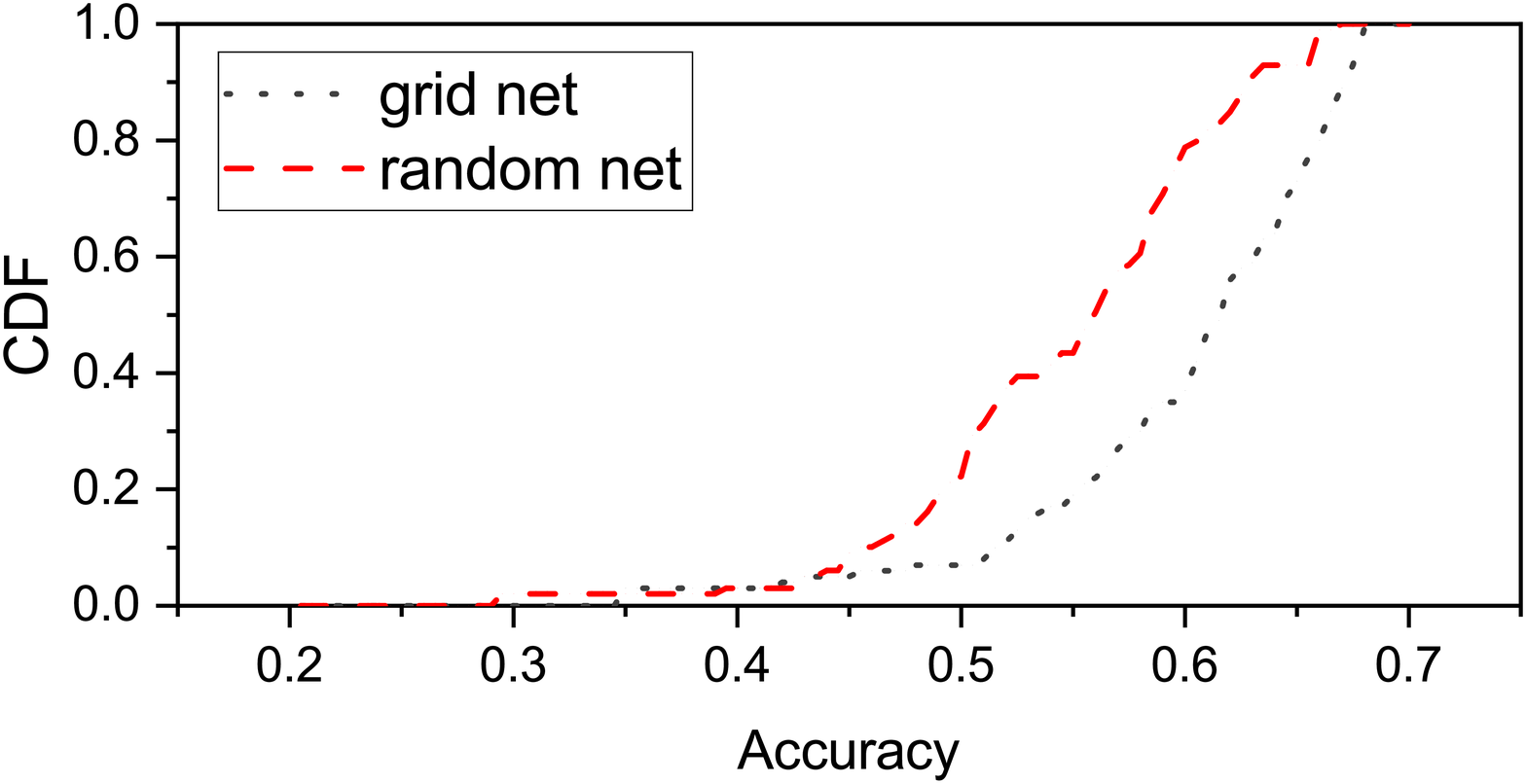}
 }
 
 \subfloat[MNIST]{
 \label{fig:Accuracy_CDF_MNIST}
 \includegraphics[width=0.37\textwidth]{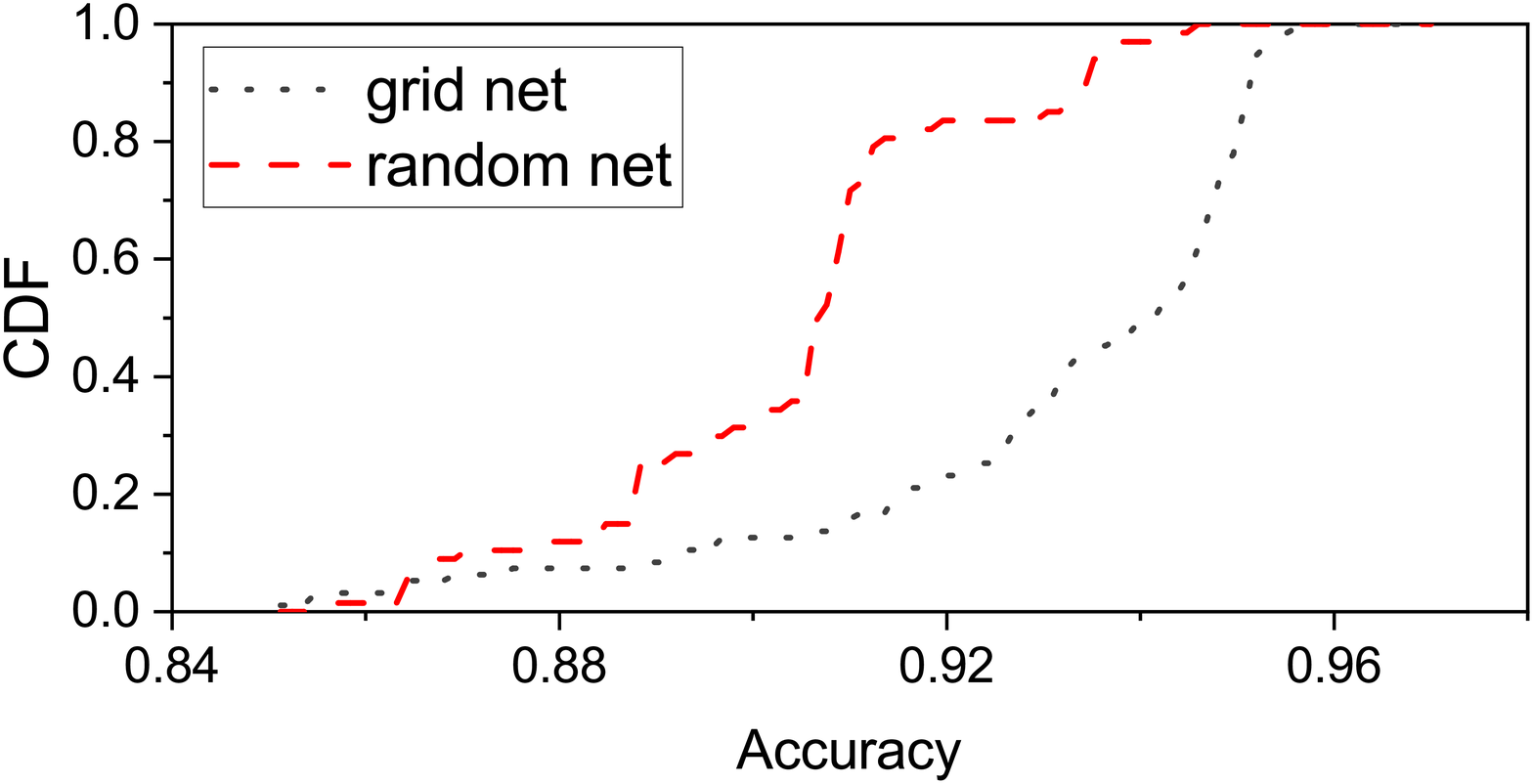}
 }
 \caption{CDFs of model accuracy of individual vehicles after conducting 3,000  iterations on CIFAR-10 and 300  iterations on MNIST in  grid and random road networks.}
 \label{fig:Accuracy_CDF_Measure}
\end{figure}

\subsection{Measuring Accuracy}

We collect the final model accuracy of each vehicle (evaluated on  test datasets) in our experiments. 
Then, we plot the cumulative distribution functions (CDF) of the final model accuracy of individual vehicles in Fig.~\ref{fig:Accuracy_CDF_Measure}. From the CDF curves of the final model accuracy, we can observe that: 1) Along vehicles' routes (randomly generated), a vehicle can only exchange model parameters with neighbour vehicles. As a result, there exist giant discrepancies of the final model accuracy on different vehicles. The accuracy can reach 70\% on CIFAR-10 or 95\% on MNIST for some lucky vehicles. In contrast,  the model accuracy can  be lower than 30\% on CIFAR-10 or 85\% on MNIST for some unlucky vehicles.   2) The network topology can severely affect the final model accuracy. The model accuracy obtained on the random topology is much worse than that on the grid topology because there exist vehicles with a low connectivity with other vehicles in random topology impeding the exchange of model parameters between vehicles.

This measurement result indicates that it is insufficient to merely consider the in-degrees and out-degrees of  vehicles to determine weights, \emph{i.e.}, $\alpha_{k,t} $'s, for decentralized model aggregation in DFL. It calls for a more sophisticated aggregation method considering data source diversity to achieve consistent high model accuracy across all vehicles in the system.

\subsection{Measuring Accuracy versus Diversity}

To explore the underlying reason resulting in poorer model accuracy for unlucky vehicles, we further investigate the relationship  between model accuracy and data source diversity. The challenge is how to measure the data source diversity since only model parameters are maintained as vehicles exchange and aggregate models with their neighbours. 
To tackle this problem, we propose to maintain state vectors on individual vehicles to track the contribution weight from each data source. The metric to measure the diversity of data sources  is based on the state vector maintained by each vehicle.

Specifically, let $\mathbf{s}_{k,t} =\{s^k_{1,t}, s^k_{2,t}, \dots, s^k_{K,t}\}$ denote the state vector of vehicle $k$ in the $t$-th global epoch where $s^k_{k',t}$  represents the contribution weight from  vehicle $k'$ to vehicle $k$'s model. 
For example, if there are $K=4$ vehicles in the system and the state vector of vehicle $1$ is $\mathbf{s}_{1,t}= \{0.25, 0.25, 0.25, 0.25\}$. It implies that the model of vehicle $1$ is equally contributed from all  vehicles. Otherwise, if $\mathbf{s}_{1,t}= \{0.50, 0.01, 0.01, 0.48\}$, it means that the model on vehicle $1$ is mainly contributed from vehicle $1$ and $4$, which  probably is not diversified enough.

Next, we describe how to update state vectors with the progress of DFL. 
Initially, all values in a state vector are assigned with $0$. State vectors are updated when vehicles conduct local iterations or model aggregations. For a particular vehicle $k$, we have 
\begin{equation}
    \label{eq:updates1}
    s_{k,t+\frac{1}{2}}^k = s_{k,t}^k + \eta_t,
\end{equation}
where $\eta_t$ is the learning rate. 
In other words, by conducting a round of local iteration, vehicle $k$ increases its own contribution weight.  Similarly, Eq.~\ref{eq:updates1} is executed for $E$ times given that local updates are conducted for $E$ rounds.  
Then, the state vector is normalized as 
\begin{equation}
    \label{eq:normalize}
    s_{k',t+\frac{1}{2}}^k  = \frac{s_{k',t+\frac{1}{2}}^k}{\sum_{k''=1}^{K}s_{k'',t+\frac{1}{2}}^k}, \quad \forall k' \in \{1, 2, \cdots, K\}
\end{equation}
By conducting a round of model aggregation, the state vector is further revised as 
\begin{equation}
    \label{eq:updates2}
    \mathbf{s}_{k,t+1} =  \sum_{k'\in\mathcal{P}_{k,t}}\alpha_{k', t}^k \mathbf{s}_{k',t+\frac{1}{2}},
\end{equation}
which is subject to $\sum_{k'\in\mathcal{P}_{k, t}}\alpha_{k', t}^k  =1$.
Here $\mathcal{P}_{k, t}$ represents the set of vehicles involved in  model aggregation conducted on vehicle $k$ in the $t$-th global iteration and $\alpha_{k', t}^k$ represents the aggregation weight assigned for the model contributed from vehicle $k'$. 

Based on 
Eqs.~\eqref{eq:updates1} and~\eqref{eq:updates2}, the functionality of state vectors can be interpreted as below. For conducting local iterations via Eq.~\eqref{eq:updates1}, a state vector incorporates the contribution of the local dataset weighted by the learning rate for $E$ rounds. 
As a vehicle aggregates its local model with other models from neighbour vehicles, its state vector is also mixed with  state vectors from  neighbours according to aggregation weights in Eq.~\eqref{eq:updates2}. In this way, a state vector can track the contribution weight from each data source.

We  embark on introducing how to measure data diversity  based on state vectors. 
We first consider a homogeneous case  to ease our explanation in which every vehicle has the same number of samples. 
It is well-known that entropy is effective in measuring diversity. Given the state vector $\mathbf{s}_{k,t}$, we define the entropy of the state vector $\mathbf{s}_{k,t}$ as
\begin{equation}
    \label{eq:entropy}
    H(\mathbf{s}_{k,t}) = -\sum_{k'=1}^{K} {s_{k',t}^k} \log_2{{s_{k',t}^k}},
\end{equation}
Apparently, $ H(\mathbf{s}_{k,t})$ achieves its maximum value $\log_2K$ if $s_{1,t}^k = \dots =s_{K,t}^k$ and minimum value $0$ if $s_{k',t}^k = 1$ for some $k'$. 

\begin{figure}[thb]
 \centering
 \subfloat[CIFAR-10]{
 \label{fig:Pearson_CIFAR-10}
 \includegraphics[width=0.37\textwidth]{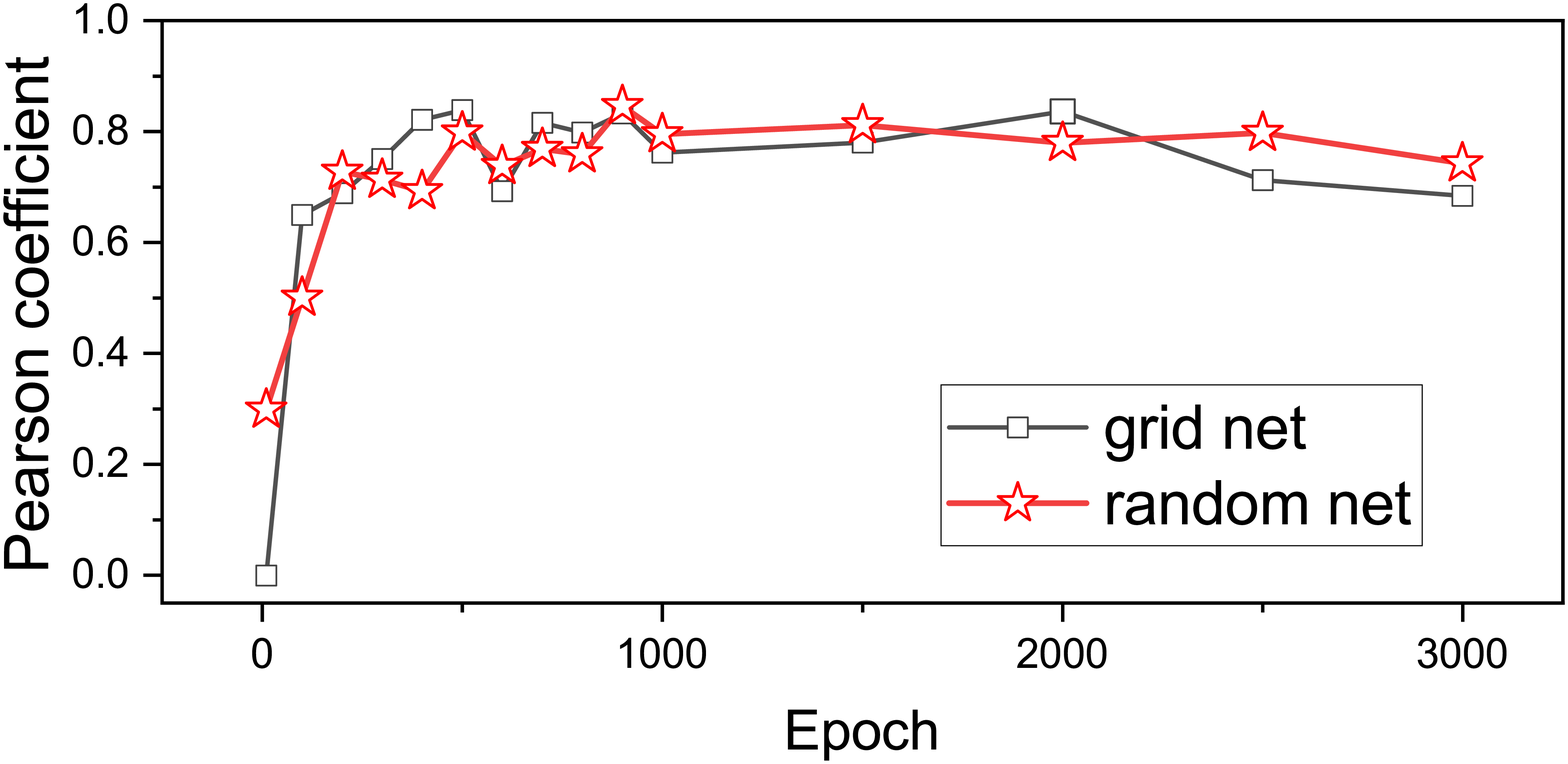}
 }
 
 \subfloat[MNIST]{
 \label{fig:Pearson_MNIST}
 \includegraphics[width=0.37\textwidth]{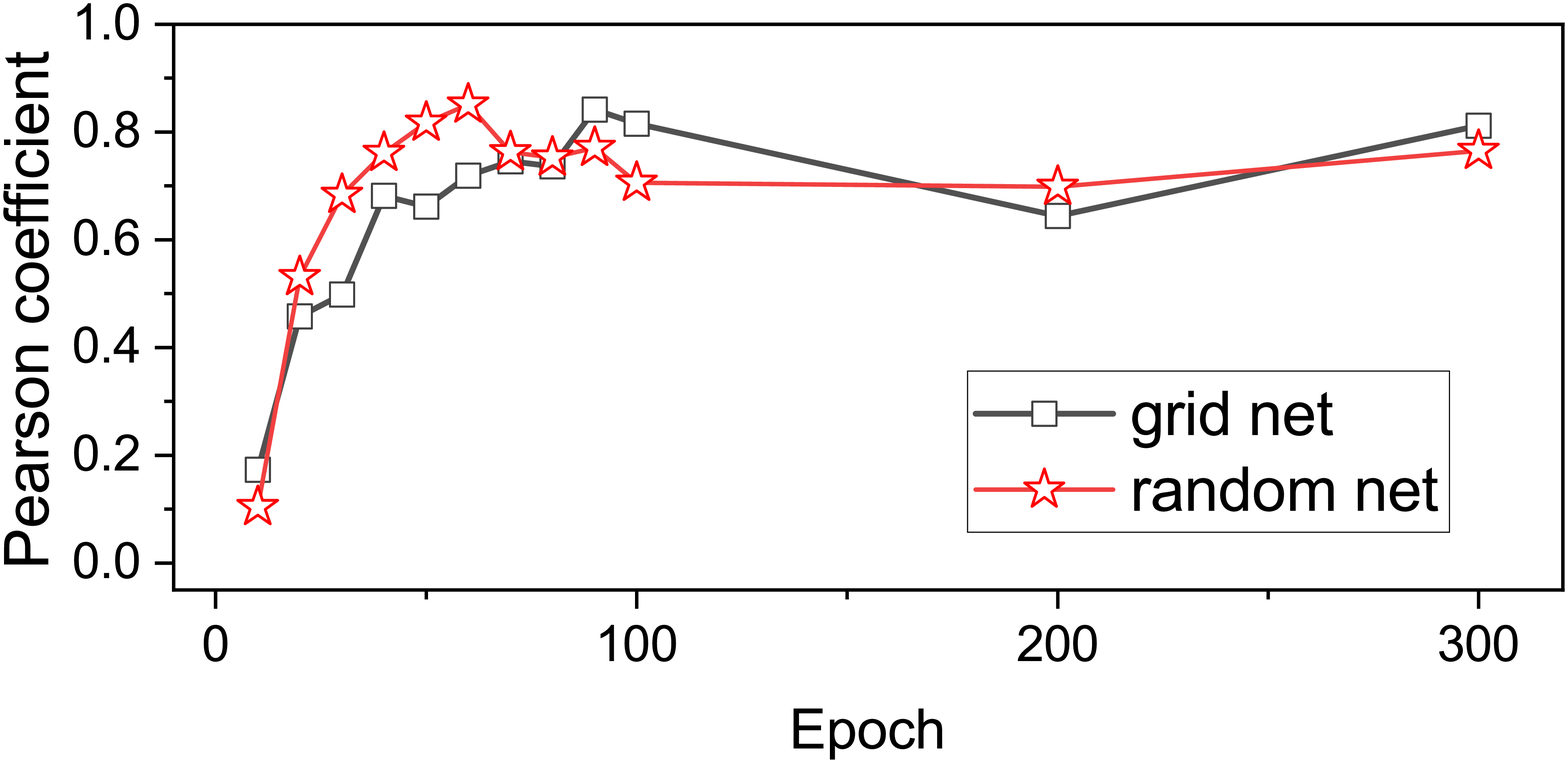}
 }
 \caption{Pearson correlation coefficients between model accuracy of individual vehicles and  entropy values of state vectors, measured based on experiments in Fig.~\ref{fig:Accuracy_CDF_Measure}. }
 \label{fig:Pearson_Measure}
\end{figure}

With the diversity measure, we can examine the importance to diversify data sources in DFL. We measure the Pearson correlation between vehicles' model accuracy and the diversity metrics of their  state vectors in the experiment presented in Fig.~\ref{fig:Pearson_Measure}. The results are plotted in Fig.~\ref{fig:Pearson_Measure}, in which the x-axis represents the index of global iteration while the y-axis represents the Pearson correlation coefficient after each  global iteration. In Fig.~\ref{fig:Pearson_Measure}, we can observe that there is a strong positive correlation between model accuracy and the diversity metric under both network topologies, indicating that unlucky vehicles with poorer model accuracy fail to diversify their data sources.
This experiment also manifests  the potential to improve model accuracy by diversifying data sources in DFL.

\section{DFL-DDS Algorithm Design}
\label{Sec:Algorithm}

This section presents the design essence of the DFL-DDS algorithm.

\subsection{Algorithm Design}

In view of the importance of data source diversity on the final model accuracy, we design the DFL-DDS algorithm that maximizes the data source diversity  when aggregating models collected from neighbour vehicles. 

The entropy value of a state vector is based on a homogeneous system. Yet,  
vehicles possibly own different sizes of datasets in practice.  To tackle the heterogeneity of  dataset  sizes  across vehicles and make our algorithm more flexible for aggregation, we propose to employ KL (Kullback–Leibler) divergence to measure data source diversity for state vectors in our algorithm.

According to \cite{FedAvg}, if the sample distribution is heterogeneous, it is common to assign the contribution weight  of a  vehicle proportional to its dataset size, \emph{i.e.}, $n_k$. To take heterogeneous  distribution into account, we define the target state vector as $\mathbf{g} = \{\frac{n_1}{n}, \frac{n_2}{n}, \dots, \frac{n_K}{n}\}$ and $g_k = \frac{n_k}{n}$, where $n = \sum_{k=1}^Kn_k$. 
The KL divergence of a state vector $\mathbf{s}_{k,t}$ is defined as its distance to the target state vector, which is 

\begin{equation}
    \label{eq:KL}
    D_{KL}(\mathbf{s}_{k,t}||\mathbf{g}) =\sum_{k'=1}^{K} {s_{k',t}^k} \log_2{\frac{s_{k',t}^k}{ g_k}}.
\end{equation}
The KL divergence achieves its minimum value when $\mathbf{s}_{k,t} = \mathbf{g}$. Intuitively speaking, the KL divergence takes both the diversity and the importance of each vehicle into account.  

\begin{figure}
\centering
\includegraphics[width=8cm]{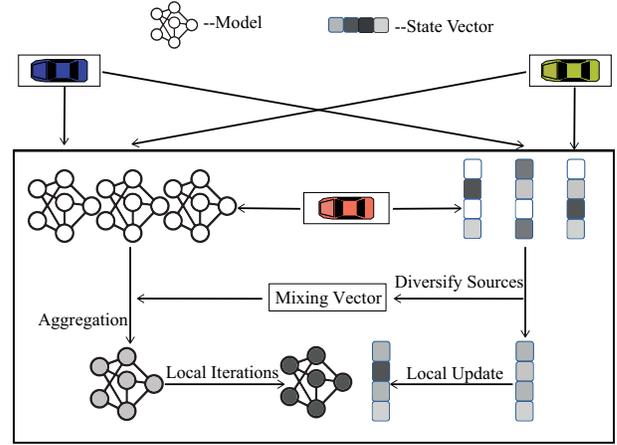}
\caption{An example to illustrate the aggregation process of DFL-DDS with the update of state vectors.}
\label{Fig:AlgFlow}
\end{figure}

Based on the defined KL divergence for each vehicle, we propose to generate weights $\alpha_{k',t}^k$ when aggregating models by minimizing the local KL divergence between the current state and the target state  in order to diversify the contribution of data sources and factor in the importance of each data source simultaneously. To facilitate the understanding of our algorithm, we exemplify the aggregation process in Fig.~\ref{Fig:AlgFlow}. In the example, it takes four steps for vehicle $k$ to update its state vector: 1) Exchange state vectors with neighbour vehicles along with the exchange of model parameters; 2) Generate weights $\alpha_{k',t}^k$'s by minimizing the KL divergence of the state vector after aggregation; 3) Aggregate models based on aggregation weight $\alpha^k_{k',t}$; 4) Update  local state vector according to Eqs.~\eqref{eq:updates1} and \eqref{eq:normalize} as local iterations are conducted. Similar to the example, the general update of state vectors can be embedded into the execution of local iterations. 
In particular, model parameters are aggregated according to the following rules in step 3). 
\begin{equation}
    \label{eq:aggregate_model}
    \mathbf{w}_{t+1}^k=\sum_{k' \in \mathcal{P}_{k,t}} \alpha^k_{k',t} \mathbf{w}_t^{k'}
\end{equation}

Note that it is not reasonable to straightly set $\alpha_{k',t}^k = g_{k'}$ for a general road network without scrutinizing the state vector of $k'$ because it overlooks the weights of data sources contributing to $\mathbf{w}_t^{k'}$. To make it clearer, we reuse the example in Fig. \ref{Fig:Case} for illustration. In Fig. \ref{Fig:Case},
there are two undirected communication paths: A-C-B and A-D. Assuming that the sample sizes in vehicles \{A, B, C, D\} are \{100, 100, 10, 100\}, respectively. Then, $\mathbf{g}=\{\frac{100}{310}, \frac{100}{310}, \frac{10}{310}, \frac{100}{310}\}$. 
However, by only considering neighbour vehicles, vehicle A will straightly set the aggregation weight of C as $\frac{10}{210}$ because it can only contact vehicles C and D without taking vehicle B into account. This example illustrates the necessity for optimizing diversity via state vectors. 
The holistic overview of our DFL-DDS algorithm is sketched in Alg.~\ref{Alg:DFL-DDS}. 

\begin{algorithm}
    \caption{Decentralized Federated Learning with Diversified Data Sources (DFL-DDS) algorithm}
    \label{Alg:DFL-DDS}
    % \hspace*{\algorithmicindent} \textbf{Input} \\
    % \hspace*{\algorithmicindent} \textbf{Output}
    \KwData{The set of vehicles, the vehicle mobility model, the road network, local epoch $E$, learning rate $\eta$}
    \KwResult{The converged models $\mathbf{w}*$}

    \SetKwBlock{DoParallel}{Do in parallel for each vehicle i}{end}
    
    {Initialize the models of all vehicles to the identical random model;}
    
    {Initialize the state vector of all vehicles;}
    
    \While{termination condition is not met for each vehicle $k$ in parallel}{
        % {XXX}
        % \Comment{XXXXX}
        % \DoParallel{
        {Receive the local model $\mathbf{w}_t^{k'}$ and state vector $\mathbf{s}_{k',t}$ from vehicle $k', \; \forall k' \in \mathcal{M}_{k,t}$;}
        
        {Minimize the KL divergence in Eq.~\eqref{eq:min_KL} to get the optimized aggregation weights $\alpha^k_{k',t}, \forall k' \in \mathcal{P}_{k,t}$;}
        
        {Aggregate all local models $\mathbf{w}_t^{k'}$, $\forall k' \in \mathcal{P}_{k,t}$, based on Eq.~\eqref{eq:aggregate_model};}
        
        {Update the aggregated model with $E$ local iterations on local training data based on Eq.~\eqref{EQ:LocalIter};}
        
        {Aggregate the state vectors of all neighboring vehicles based on Eq.~\eqref{eq:updates2};}

        {Update aggregated state vector based on Eq.~\eqref{eq:updates1};}
        
        {Normalize the state vector based on Eq.~\eqref{eq:normalize};}
        % }
    }
\end{algorithm}

\subsection{Analysis}

To justify the feasibility of step 2) deducing $\alpha_{k, t}$ based on state vectors, we move on to prove that minimizing KL divergence can be solved efficiently by vehicles.

According to Eq.~\eqref{eq:updates2}, the problem to minimize KL divergence is formulated as
\begin{equation}
\label{eq:min_KL}
\begin{aligned}
               \mathbb{P}1:     & \min_{  \alpha_{k',t}^k, \forall k'\in \mathcal{P}_{k,t}  } D_{KL}(\mathbf{s}_{k,t+1}||\mathbf{g})       \\
    s.t. \quad      & \eqref{eq:updates2} \\
                    & \sum_{k' \in \mathcal{P}_{k,t}} \alpha_{k', t}^k = 1,         \\
                    & 0\leq \alpha_{k',t}^k \leq 1, \quad \forall k' \in \mathcal{P}_{k,t},    \\
                    & \alpha_{k',t}^k = 0, \quad\forall k' \notin \mathcal{P}_{k,t}.    \\
\end{aligned}
\end{equation}
It is known that the KL divergence as the objective of $\mathbb{P}1$ is a convex function. Meanwhile, all constraints in $\mathbb{P}1$ are linear functions. Hence, this is a convex optimization problem, which can be solved efficiently to optimally determine weights  $\alpha_{k',t}^k$'s for model aggregation.

It is worth noting that the problem $\mathbb{P}1$ minimizing the KL divergence of the local state vector is equivalent to maximizing the entropy value of the state vector when the sample distribution is balanced, \emph{i.e.}, $\mathbf{g} = \{\frac{1}{K},\dots, \frac{1}{K}\}$.
For this special case, we have 
$ D_{KL}(\mathbf{s}_{k,t}||\mathbf{g}) =\sum_{k'=1}^{K} {s_{k',t}^k} \log_2{{s_{k',t}^k}}-\log_2 \frac{1}{K} = -H(\mathbf{s}_{k,t})+\log_2 {K}.$

\subsection{Implementation}
DFL-DDS is friendly for implementation in practice. 
Firstly, the computation and communication overhead incurred by DFL-DDS is negligible. Given $K$ vehicles in the system, exchanging state vectors only brings $O(K)$ communication overhead, which could be much less than the communication load for exchanging complicated CNN models between vehicles. 
The computation overhead lies in solving problem $\mathbb{P}1$, a small-scale convex optimization problem. It is still insignificant compared with the computation load to train CNN models.  
Secondly, the communication overhead can be further reduced if the system scale $K$ is large by maintaining dynamic state vectors on vehicles.  In other words, each vehicle can adaptively adjust the size of its state vector $\mathbf{s}_{k, t}$ to only record weights from vehicles which have already contributed to its model training. Obtaining and distributing the target vector may consume extra communication traffic as well. Yet, the target state vector is static and limited with $O(K)$. It implies that a central coordinator can easily complete this task with low overhead.

Besides, our algorithm can be easily extended if RSUs are available for supporting DFL. An RSU can be regarded as a special static vehicle, which can also maintain a state vector to record the contribution weight of each vehicle. Thus, a vehicle can exchange information with an RSU to accelerate DFL as long as they are within effective communication distance.

\section{Experimental Results and Analysis}
\label{Sec:Experiment}

In this section, we report the results of  experiments conducted with MNIST and CIFAR-10 datasets to evaluate our DFL-DDS algorithm. 

\subsection{Experimental Settings}

% CNN model
\subsubsection{Datasets}
We employ the well-known image datasets CIFAR-10 and MNIST for our experiments, since they are widely used in many areas including vehicular networks \cite{Chai2021, nishio2019client}. The CIFAR-10 dataset consists of 50,000 training samples and 10,000 testing samples. Each sample in the CIFAR-10 dataset is a 3*32*32 color image  with one of ten labels such as ships, cats, and dogs. The MNIST dataset consists of 60,000 training samples, and 10,000 testing samples. Each sample is a 28*28 handwritten digit with a label of digits from 0 to 9.

\subsubsection{Learning Tasks}

The learning tasks on each vehicle are to train two CNN (Convolutional Neural Network) models for  classifing CIFAR-10 and MNIST images, respectively.  The structure of two CNN models are described as below:\footnote{These models are from the public code repository at https://github.com/AshwinRJ/Federated-Learning-PyTorch}
\begin{itemize}
    \item The CNN to classify CIFAR-10 images is  with three 3$\times$3 convolution layers (the first layer is with 16 channels, the second one is with 32, the third one is with 64 and each channel is followed by a 2$\times$2 max pooling), a dropout layer with the probability of 0.25 and a fully connected layer with a log softmax output layer. There are 33,834  parameters in total.
    \item The  model to classify MNIST images is with two 5$\times$5 convolution layers (the first layer is with 10 channels, the second one is with 20 channels and each channel is followed with 2$\times$2 max pooling), a fully connected layer with 50 units, a dropout layer with the probability of 0.5, a final fully connected layer with a log softmax output layer. There are 21,840 parameters in total.
\end{itemize}

\subsubsection{Road Network and Trajectory Generation}

\begin{figure}[htb]
 \centering
 \subfloat[Grid net]{
 \label{fig:grid_net}
 \includegraphics[width=0.1\textwidth]{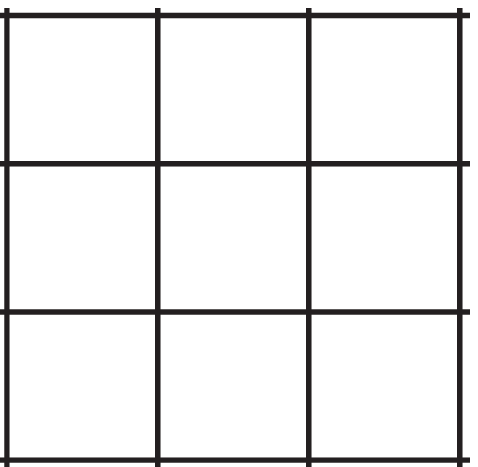}
 }
 \subfloat[Random net]{
 \label{fig:random_net}
 \includegraphics[width=0.1\textwidth]{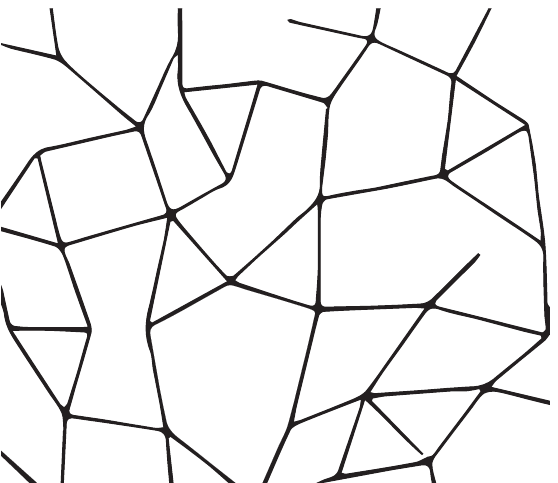}
 }
 \subfloat[Spider net]{
 \label{fig:spider_net}
 \includegraphics[width=0.12\textwidth]{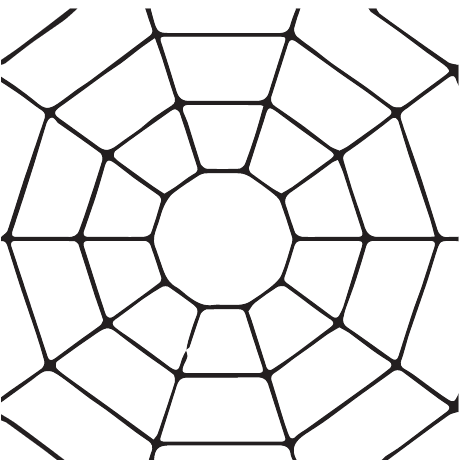}
 }
 \caption{Three types of road network topologies.}
 \label{fig:road_topology}
\end{figure}

We generate three road networks for our experiments. Road network examples are shown in Fig.~\ref{fig:road_topology}.
\begin{itemize}
    \item Grid net: We set the number of nodes as 10 for both horizontal  and vertical directions, and the length between two neighboring nodes is  100 meters. As a result, 100 nodes are placed evenly in the grid network, and their degrees range from 2 to 4 with frequencies \{4, 32, 64\}. 
    \item Random net: The distance between two neighboring nodes is randomly generated in the range from 100 to 200 meters and other parameters are the same as the default values in the SUMO simulator \cite{SUMO2018}. In the random network, 100 nodes  are randomly generated  with 100 iterations. For these nodes, their degrees range from 1 to 5 with frequencies \{25, 7, 36, 27, 5\}. 
    \item Spider net: The number of nodes and the distance between nodes are determined by three parameters:  arms, circles, and radius increment of neighboring circles, which are set as 10, 10, and 100 meters, respectively. We locate 100 nodes evenly in 10 circles.
\end{itemize}

We adopt the same settings as the simulation study for the generation and communication of vehicles. There are 100 vehicles moving in each road network.  Their trajectories are generated based on the Manhattan mobility model \cite{Manhattan}. The default velocity of each vehicle is 13.89 m/s, while the real velocity is influenced by several factors, such as road types and the congestion condition. We assume that all vehicles are equipped with DSRC and mmWave communication components with a communication range of 100 meters\cite{FVN, cao2021, zhao2020reinforcement}.

\begin{table}
    \centering
    \caption{Important parameters in experiments.}
    \label{Tab:hyperparameter}
    \begin{tabular}{c|c}
        \hline
        Experiment Parameter                & Value              \\
        \hline
        learning rate $\eta$   & 0.1           \\
        local minibatch size $B$      & 80            \\
        local epoch $E$     & 8             \\
        vehicle number $K$  & 100           \\
        communication range $r$ & 100m      \\
        \hline
    \end{tabular}
\end{table}

For convenience, we summarize critical parameters in our experiments in Table~\ref{Tab:hyperparameter}.
\subsubsection{Sample Distribution}
% data distribution
In FL, the typical data sample distribution is non-IID (not independent and identically distributed) and unbalanced. Accordingly, we adopt two different ways to place images on vehicles for our experiments.
\begin{itemize}
    \item Balanced \& non-IID: The datasets are firstly grouped based on their labels, which are then evenly partitioned into $4 *K$ shards. Each vehicle is assigned with 4 shards so that each vehicle receives the same number of training samples. The number of labels owned by each vehicle is from 2 to 4. 
    \item Unbalanced \& IID: All samples assigned to a vehicle are uniformly and randomly selected from CIFAR-10 and MNIST in an IID manner. But, the number of samples on each vehicle is restricted to one of values in \{125, 375, 1125\} for CIFAR-10 or \{150, 450, 1350\} for MNIST.    
    
\end{itemize}

\subsubsection{Baselines and Evaluation Metrics}

We implement two baselines:  the subgraident push (SP) algorithm proposed in \cite{TimeVaryingGraph} and the decentralized federated learning (DFL) algorithm proposed in \cite{Savazzi_20}.
In the SP algorithm,  each vehicle only conducts local iteration with all local samples once before it evenly partitions its  parameters and broadcasts them to all neighbor vehicles. The DFL algorithm is a distributed version of the FedAvg algorithm, which regards each vehicle as a PS to  execute FedAvg. Its parameters are set as the same values as our DFL-DDS except the model aggregation weights. For DFL, it assigns weights to  models proportional to the sample population on each involved vehicle. 

We evaluate DFL-DDS and baselines with three metrics: 1) The overall average model accuracy; 2) The number of consumed global epochs to reach target accuracy; 3) Consensus distance which is defined as $\Xi^2_t = \frac{1}{K} \sum_{k=1}^{K}||\bar{\mathbf{w}}_{t} - \mathbf{w}^k_t ||^2$.  Here $\bar{\mathbf{w}}_{t} = \frac{1}{K} \sum_{k=1}^{K} \mathbf{w}^k_t$, and thus $\Xi^2_t$ represents the average distance between the virtual global model and local models.

If all samples are centrally located on a server, the accuracy of our CNN models on CIFAR-10 and MNIST datasets can reach about 72\% and 98\%, respectively. For our experiments,  the final average model accuracy is comparable with that obtained with centralized training.

\subsection{Experimental Results}

\begin{figure}
\centering
\includegraphics[width=0.8\linewidth]{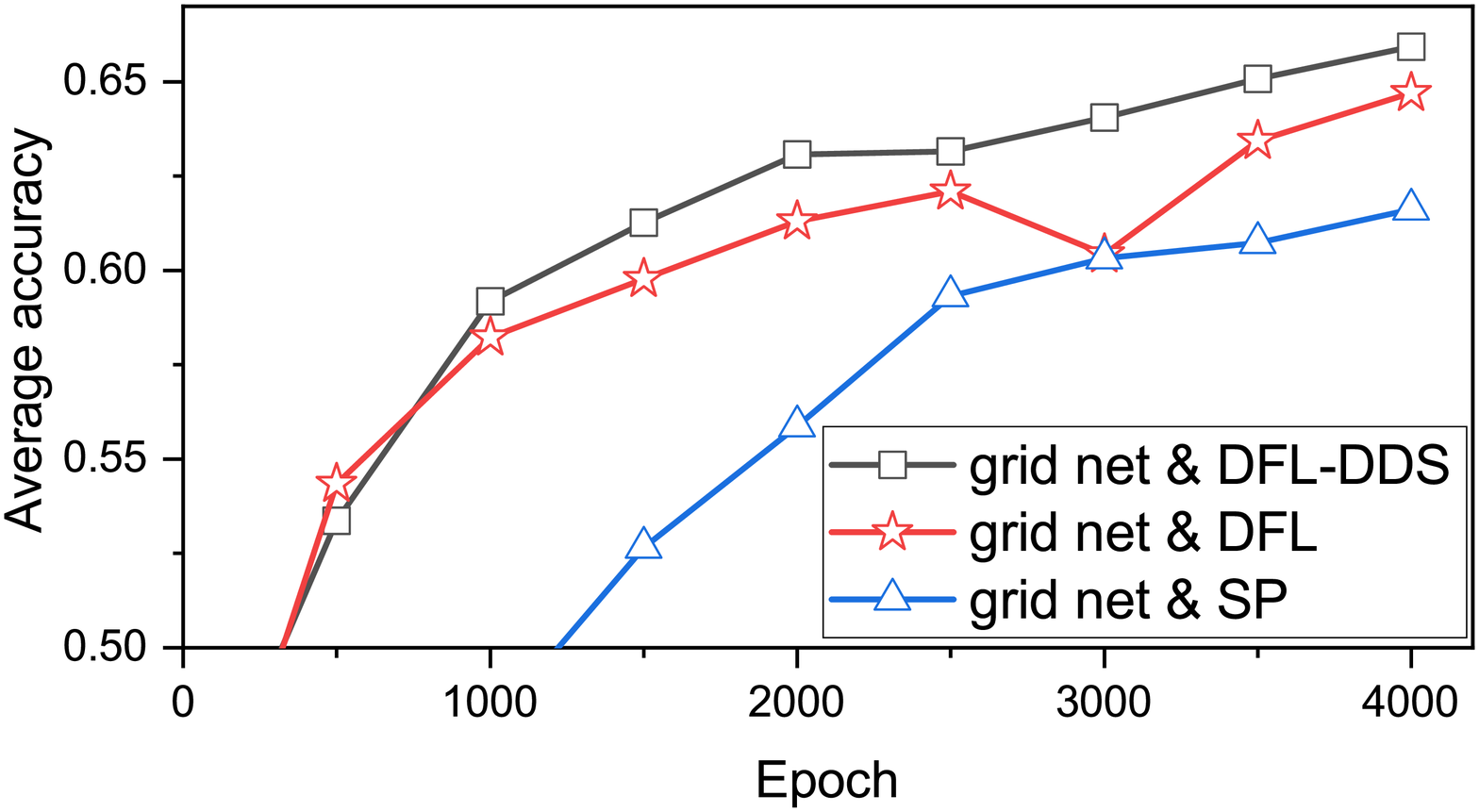}
\caption{The average accuracy of 100 vehicles at each training epoch with CIFAR-10. (Balanced \& non-IID)}
\label{fig:accuracy-epoch-CIFAR-10}
\end{figure}

\begin{figure}
\centering
\includegraphics[width=0.8\linewidth]{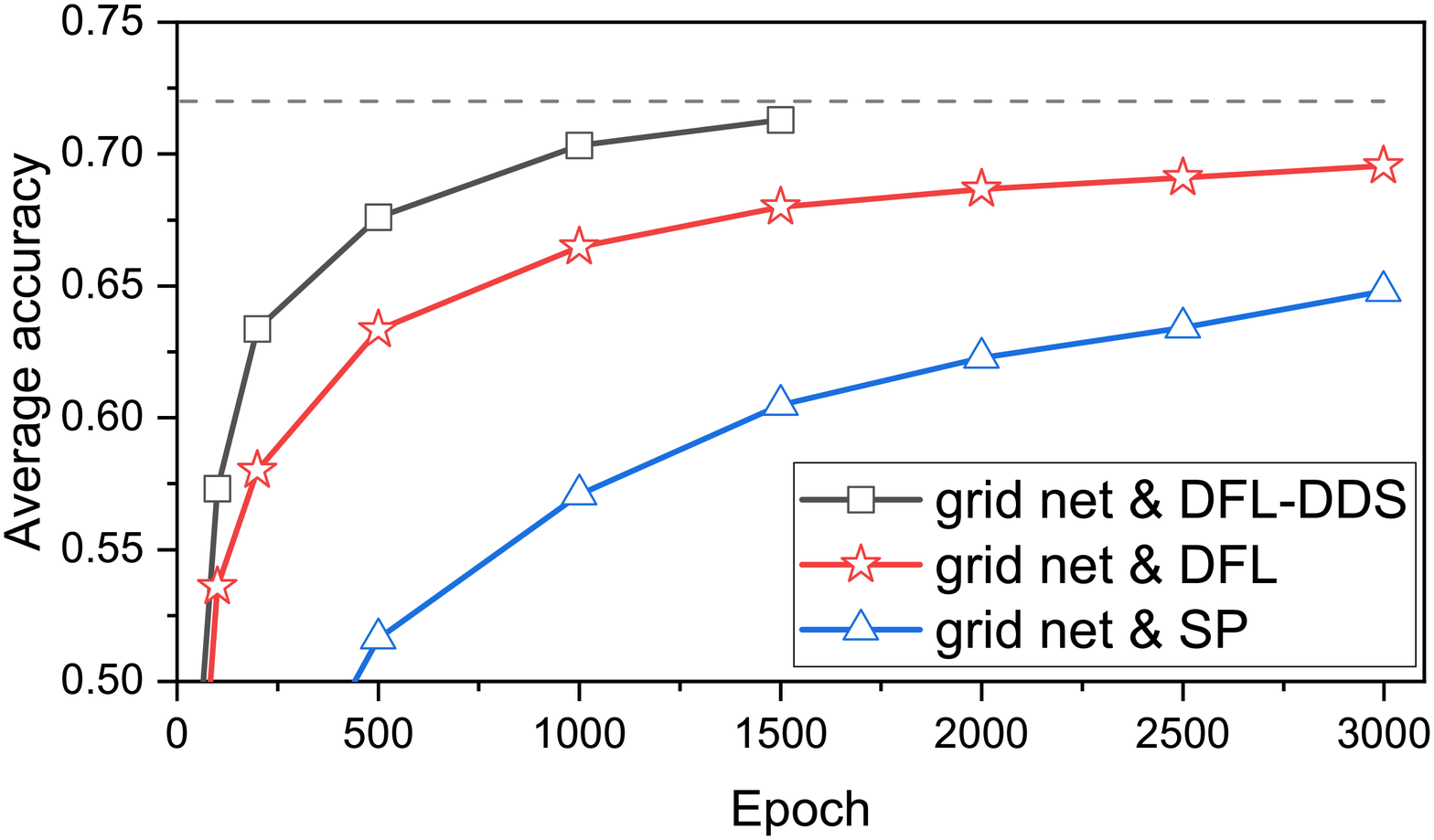}
\caption{The average accuracy of 100 vehicles at each training epoch with CIFAR-10 (Unbalanced \& IID).}
\label{fig:g-8}
\end{figure}

\subsubsection{Evaluating Accuracy with CIFAR-10}
In Figs.~\ref{fig:accuracy-epoch-CIFAR-10} and \ref{fig:g-8}, we conduct experiments using the CIFAR-10 dataset to  compare the model accuracy of DFL-DDS, DFL and SP under two different dataset distribution scenarios and the grid road network. Due to the complication of the CIFAR-10 dataset, we respectively conduct 4,000 and 3,000 global epochs in total under two dataset distributions.   The x-axis represents the number of conducted global epochs, while the y-axis represents the average model accuracy of 100 vehicles after each epoch. 
From the results in Figs.~\ref{fig:accuracy-epoch-CIFAR-10} and \ref{fig:g-8}, we can observe that:
\begin{itemize}
    \item The DFL-DDS algorithm outperforms  DFL and SP in terms of average model accuracy under both sample distribution scenarios. For instance, after 2,000 epochs, the average accuracy of DFL-DDS reaches 63\%, while the DFL and SP algorithms only reach 61\% and 56\%, respectively, under the non-IID sample distribution scenario.
    \item The data sample distribution substantially influences the final model accuracy. For the non-IID sample distribution scenario, the final model accuracy is less than 70\%. Whereas, for the IID scenario, the model accuracy of DFL-DDS exceeds 71\% after about 1,500 global epochs, which is significantly higher than that of DFL and SP. This accuracy is already about the same as that achieved with centralized training, and thereby the execution of DFL-DDS is terminated then. 
    \item For the Unbalanced \& IID distribution, vehicles may possess different numbers of samples. The results in Fig.~\ref{fig:g-8} confirm that it is reasonable to adopt KL divergence to measure the diversity of data sources, which can well handle the heterogeneity of sample sizes. 
\end{itemize}

\subsubsection{Evaluating Accuracy with MNIST}

\begin{figure}[htb]
 \centering
 \subfloat[Grid net]{
 \label{fig:g-1}
 \includegraphics[width=0.34\textwidth]{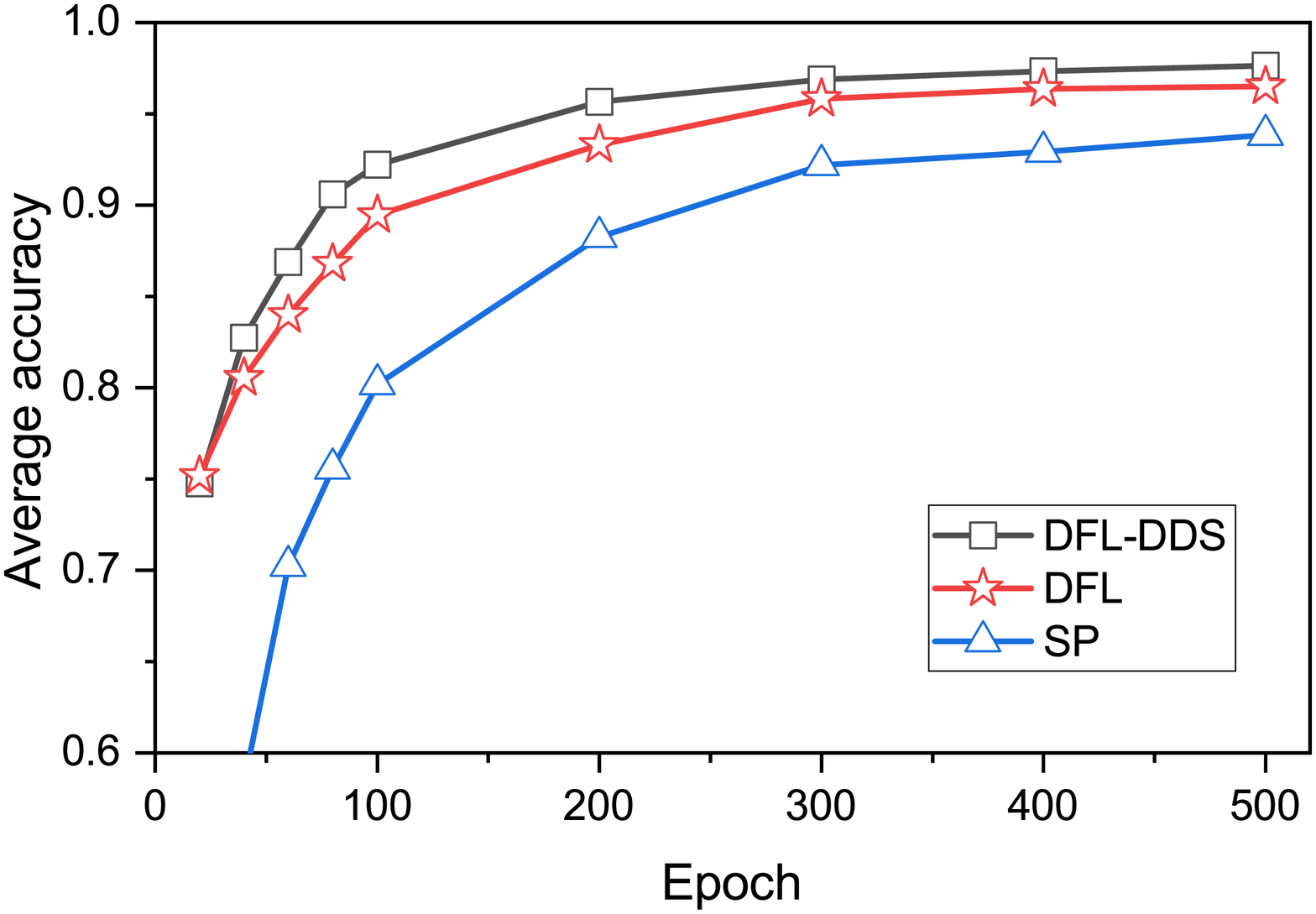}
 }
 
 \subfloat[Random net]{
 \label{fig:g-2}
 \includegraphics[width=0.34\textwidth]{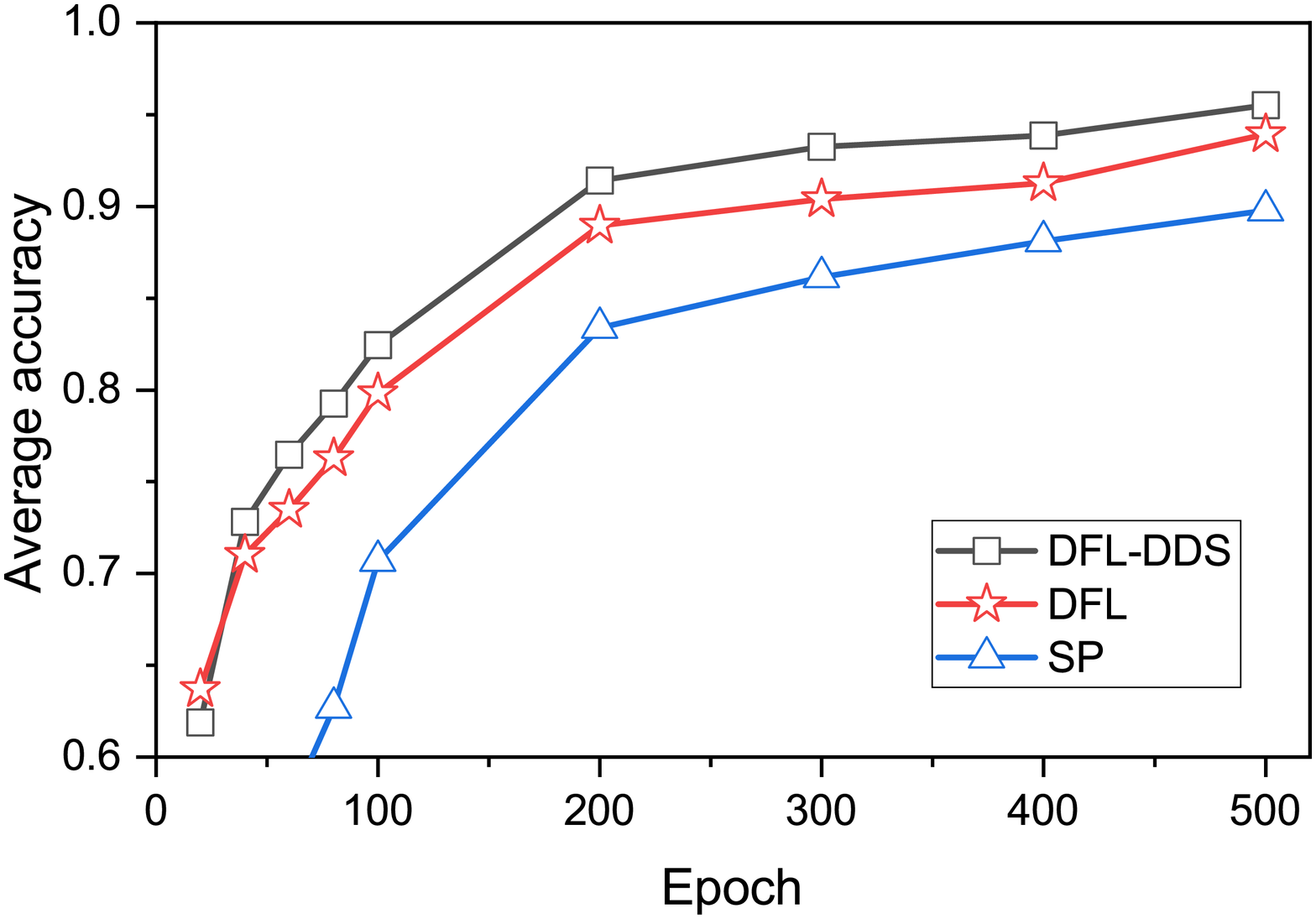}
 }
 
  \subfloat[Spider net]{
 \label{fig:g-3}
 \includegraphics[width=0.34\textwidth]{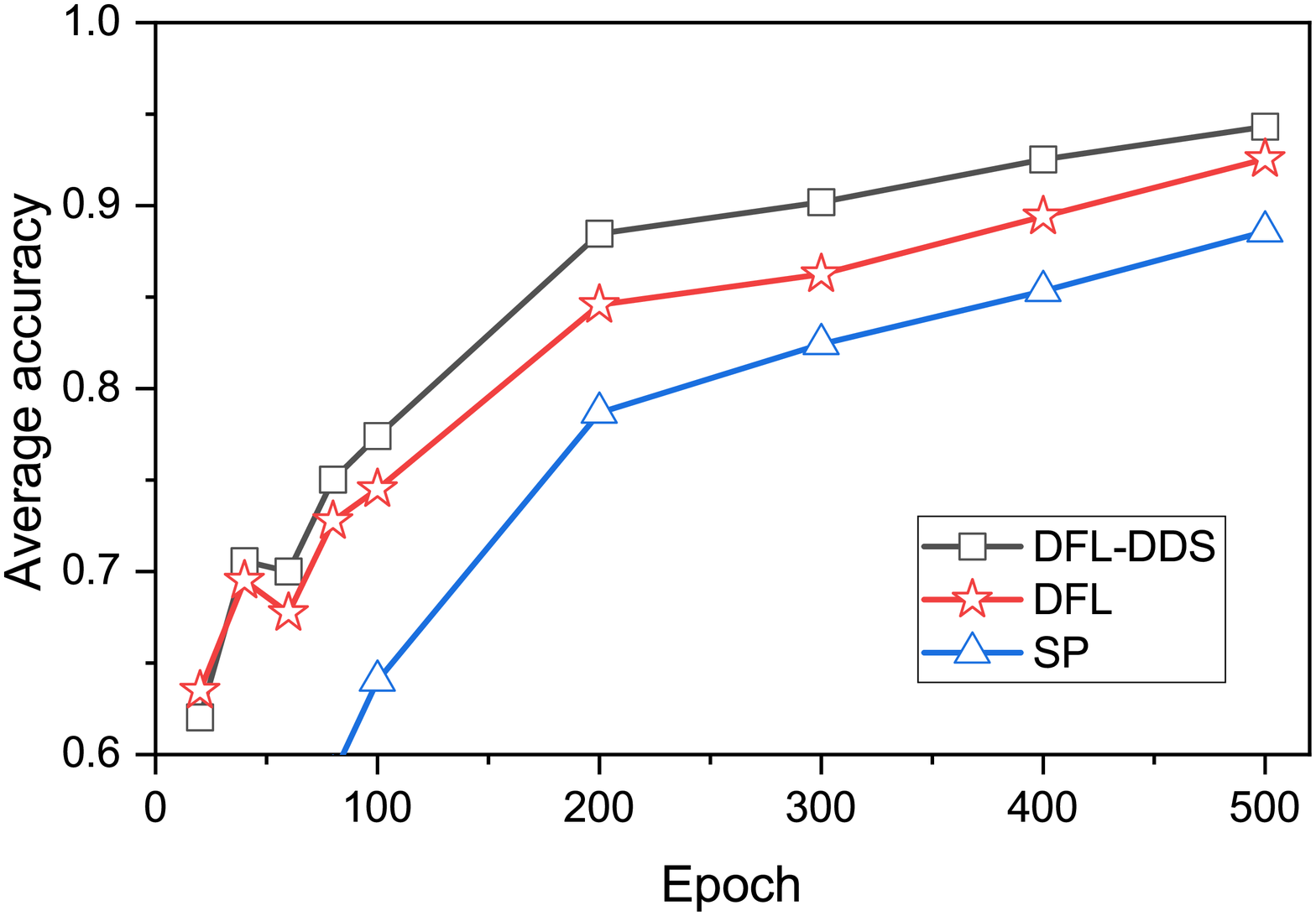}
 }
 
 \caption{The average accuracy of 100 vehicles at each training epoch with MNIST. (Balanced \& non-IID)}
 \label{fig:g_average}
\end{figure}

We further evaluate the performance of DFL-DDS and two baselines, \emph{i.e.}, DFL and SP,  using the MNIST dataset under three kinds of road networks. Due to the simplicity of the MNIST dataset, we totally execute 500 global epochs on each vehicle for all three algorithms. The experiment results are presented in Fig.~\ref{fig:g_average}. The x-axis represents the number of conducted global epochs while the y-axis represents the average model accuracy of 100 vehicles on the test dataset. From Fig.~\ref{fig:g_average}, we can observe that:
\begin{itemize}
    \item DFL-DDS is the best one that achieves the highest model accuracy under all evaluation cases. The model accuracy of DFL-DDS can finally exceed 95\% which is very close to the accuracy achieved by centralized training. The results of this experiment  validate the superiority of DFL-DDS by diversifying data sources for model aggregation.  
    \item
    DFL is better than SP (which only conducts  local iteration once per global epoch with the full set of local samples) because DFL conducts multiple local iterations with local sample batches per global epoch which can make the trained model converge with a faster rate.   
    \item The network topology can substantially affect the model accuracy. The model accuracy in the grid network is the best one while the accuracy in the spider network is the worst one.  The reason is that both the degrees of junctions and the lengths of each road segments are uniform in grid network implying a high connectivity among vehicles. In the random  network, some junctions have a low degree, which leads to a low connectivity for some unlucky vehicles. In the  spider network, the perimeters of different circles are highly different, and vehicles in outer circles have sparser connectivity with other vehicles.  Overall speaking,  if the road network is more irregular and  sparser, it is possible that more vehicles will have difficulties to diversify their data sources resulting in lower model accuracy. 
\end{itemize}

\begin{figure}[t]
\centering
\includegraphics[width=0.95\linewidth]{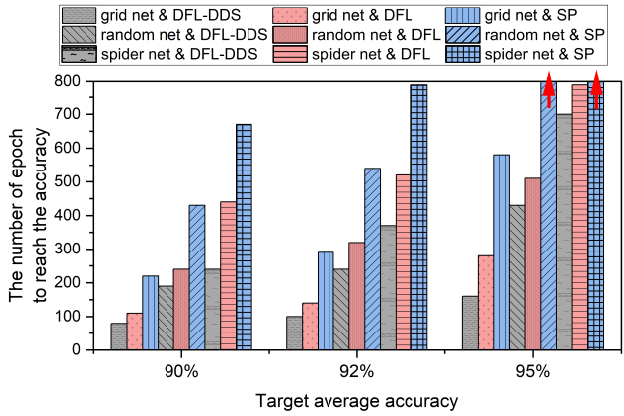}
\caption{The number of epochs to reach 90\%, 92\% and 95\% with MNIST (Balanced \& non-IID).}
\label{fig:how-many-epoch}
\end{figure}

\subsubsection{Evaluating Iteration Numbers}
From another perspective, we compare DFL-DDS with baselines in terms of the number of global iterations consumed to reach the target average model accuracy. From this comparison, we can investigate how much time cost can be reduced by diversifying data sources in DFL. We employ the MNIST dataset and enumerate the target average model accuracy as 90\%, 92\% and 95\%. The required numbers of global epochs for each experiment case are displayed in Fig.~\ref{fig:how-many-epoch}. Our experiment results reveal that:
\begin{itemize}
    \item DFL-DDS is the best one always taking the fewest number of global epochs to reach the target model accuracy in all experiment cases. Note that there are two experiment cases in which some algorithms cannot reach the target average model accuracy. These two cases are marked by red arrows in Fig.~\ref{fig:how-many-epoch}.
    \item Compared with SP, the DFL-DDS algorithm can reduce the number of epochs to reach the target by more than 53\%. In contrast, the reduction of epochs compared with the DFL algorithm varies from about 11\% to 45\%. These results  shed light on the merit of DFL-DDS which can considerably improve the training efficiency of DFL. 
\end{itemize}

\subsubsection{Evaluating Consensus Distance}
The consensus among participating clients is pivotal in evaluating the convergence speed of decentralized machine learning. Previous works have proposed several similar metrics such as consensus distance \cite{kong2021consensus} and Laplacian potential \cite{Hongli2022} to measure the degree of disagreement among all local models. 
Inspired by these works, we employ consensus distance defined as $\Xi^2_t$ to evaluate DFL progress in our experiments. A lower consensus distance implies a better DFL performance.

\begin{figure}[t]
 \centering
 \subfloat[CIFAR-10+IID]{
 \label{fig:consensus-cifar-10}
 \includegraphics[width=0.4\textwidth]{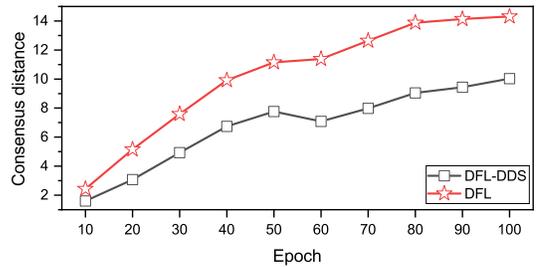}
 }
 
 \subfloat[MNIST+Non-IID]{
 \label{fig:consensus-mnist}
 \includegraphics[width=0.4\textwidth]{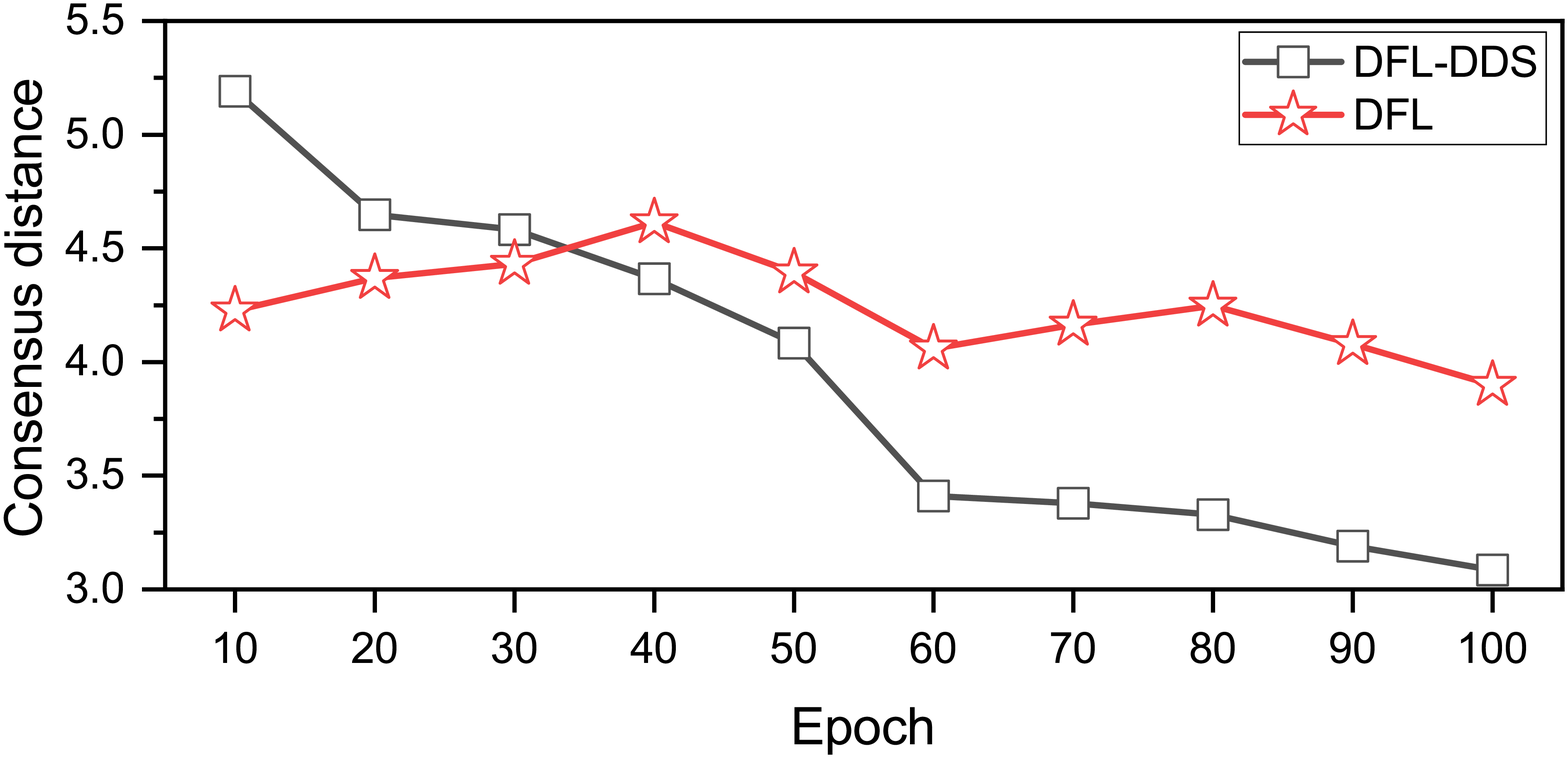}
 }
 \caption{The comparison of consensus distance between DFL-DDS and DFL for the first 100  global epochs.}
 \label{fig:consensus-distance}
\end{figure}

Similar to \cite{kong2021consensus}, we conduct the experiment
to evaluate the consensus distances of DFL-DDS and DFL for the first 100 global epochs. We set up two experiment cases: IID dataset distribution with CIFAR-10 and non-IID dataset distribution with MNIST. The road network is generated as a grid net. Fig.~\ref{fig:consensus-distance} displays the comparison of  the consensus distance for the first 100 epochs. From results in Fig.~\ref{fig:consensus-distance}, we can observe that the consensus distance of DFL-DDS is always much lower than that of  DFL, indicating that DFL-DDS can significantly outperform DFL by diversifying data sources.

\section{Conclusion and Future Work}
\label{Sec:Con}
 DFL is naturally feasible for vehicular networks.
However, the low training efficiency is the major obstacle impeding the wide application of DFL.
Through a simulation study, we unveil that the lack of data source diversity is a main reason resulting in poor model accuracy. 
To address this problem, we propose a novel DFL-DDS algorithm  by maximizing the data source diversity contributed to the model trained on individual vehicles. 
Meanwhile, we elaborate  implementation issues of DFL-DDS by considering practical issues. 
Experimental results conducted with public datasets confirm the superiority of DFL-DDS. 

This work as our initial attempt to improve DFL is insightful for future works in this area. Our future work includes two possible aspects. 
Firstly, communication unreliability has not been considered in our algorithm. In other words, vehicle communications can be interrupted due to various reasons such that models cannot be completely exchanged between vehicles. Robust strategies should be developed that can tolerate information loss. 
Secondly, we assume a static target state vector in our work. In practice, the sample population on vehicles can be a dynamic variable. A dynamic lightweight algorithm is needed to timely adjust the target state vector. 

\clearpage
%\balance
\bibliographystyle{IEEEtran}
\small
\bibliography{main_bib}

\end{document}